\documentclass[conference]{IEEEtran}
\IEEEoverridecommandlockouts
\usepackage{cite}
\usepackage{amsmath,amssymb,amsfonts}
\usepackage{algorithmic}
\usepackage{graphicx}
\usepackage{textcomp}
\usepackage{xcolor}
\usepackage{multirow}
\usepackage{hyperref}

\def\BibTeX{{\rm B\kern-.05em{\sc i\kern-.025em b}\kern-.08em
    T\kern-.1667em\lower.7ex\hbox{E}\kern-.125emX}}

\newcommand{\fakeparagraph}[1]{\vspace{1mm}\noindent\textbf{#1}}
    
\begin{document}

\title{BiSeg-SAM: Weakly-Supervised Post-Processing Framework for Boosting Binary Segmentation in Segment Anything Models\\
{\footnotesize \textsuperscript{}}
\thanks{$^*$Hu Cao is the corresponding author of this work (hu.cao@tum.de)}
}

\author{\IEEEauthorblockN{1\textsuperscript{st} Encheng Su}
\IEEEauthorblockA{\textit{Engineering and Design} \\
\textit{Technische Universität München}\\
München, Germany \\
encgo.su@tum.de}
\and
\IEEEauthorblockN{2\textsuperscript{nd} Hu Cao$^*$}
\IEEEauthorblockA{\textit{Computation, Information and Technology} \\
\textit{Technische Universität München}\\
München, Germany \\
hu.cao@tum.de}
\and
\IEEEauthorblockN{3\textsuperscript{rd} Alois Knoll}
\IEEEauthorblockA{\textit{Computation, Information and Technology } \\
\textit{Technische Universität München}\\
München, Germany \\
knoll@in.tum.de}
}

\maketitle

\begin{abstract}
Accurate segmentation of polyps and skin lesions is essential for diagnosing colorectal and skin cancers. While various segmentation methods for polyps and skin lesions using fully supervised deep learning techniques have been developed, the pixel-level annotation of medical images by doctors is both time-consuming and costly. Foundational vision models like the Segment Anything Model (SAM) have demonstrated superior performance; however, directly applying SAM to medical segmentation may not yield satisfactory results due to the lack of domain-specific medical knowledge. In this paper, we propose BiSeg-SAM, a SAM-guided weakly supervised prompting and boundary refinement network for the segmentation of polyps and skin lesions. Specifically, we fine-tune SAM combined with a CNN module to learn local features. We introduce a WeakBox with two functions: automatically generating box prompts for the SAM model and using our proposed Multi-choice Mask-to-Box (MM2B) transformation for rough mask-to-box conversion, addressing the mismatch between coarse labels and precise predictions. Additionally, we apply scale consistency (SC) loss for prediction scale alignment. Our DetailRefine module enhances boundary precision and segmentation accuracy by refining coarse predictions using a limited amount of ground truth labels. This comprehensive approach enables BiSeg-SAM to achieve excellent multi-task segmentation performance. Our method demonstrates significant superiority over state-of-the-art (SOTA) methods when tested on five polyp datasets and one skin cancer dataset. The code for this work is open-sourced and available at https://github.com/suencgo/BiSeg-SAM.
\end{abstract}


\section{Introduction}
Accurate segmentation is essential for early diagnosis and treatment planning in medical imaging, particularly for colorectal cancer (CRC) and skin cancer. Automated segmentation methods significantly enhance the accuracy and efficiency of lesion detection. CRC has a high global incidence and mortality rate~\cite{b1}, while melanoma, a type of skin cancer, poses a major threat to public health. These challenges underscore the need for precise lesion identification to enable early diagnosis and effective treatment.

Convolutional Neural Networks (CNNs), such as U-Net~\cite{b18}, MSNet~\cite{b14}, CaraNet~\cite{b15}, UNet++~\cite{b4}, and UACANet~\cite{b16}, have brought substantial advancements to segmentation. Their ability to learn complex features has greatly enhanced segmentation accuracy and reliability. Transformer-based models have achieved significant progress in the field of computer vision~\cite{han2022survey, GhostViT, SDPT}. In medical image segmentation, many excellent methods are proposed, such as Polyp-PVT~\cite{b5} and Swin-Unet~\cite{b20}. However, these methods heavily rely on large, precisely annotated datasets, which are costly and time-consuming to acquire. The advent of large models like SAM~\cite{b10} and its variants, including polyp-SAM~\cite{b12}, Poly-SAM++\cite{b13}, and BA-SAM\cite{song2024ba}, has set new benchmarks in performance. Yet, the high computational demands of these models and the uncertainty of outcomes highlight the ongoing challenge of requiring high-quality data.

To address these limitations, we introduce BiSeg-SAM, a weakly supervised learning network tailored for SAM, optimizing segmentation tasks with limited annotations for binary classification in colorectal and skin cancer imaging. Unlike prior works that employed scribbles for supervision, our approach utilizes weak box supervision within SAM for medical image segmentation. We enhance the weak box supervision method by adapting bounding boxes based on the foreground center points, which has shown promising results. Our Adaptively Global-Local Module incorporates a CNN block to capture local features, improving the model's ability to process fine details in the image. Additionally, our DetailRefine Module refines coarse segmentation predictions, thereby improving boundary precision and overall segmentation accuracy.

While weak supervision reduces the labeling costs by allowing the use of imprecisely labeled data, during testing, we observed that the bounding box generation method, M2B in WeaklyPolyp~\cite{b9}, introduced excessive noise when handling multiple foreground objects. This often resulted in large non-foreground areas being included, leading to suboptimal performance on complex binary classification data.

\begin{figure}[t]
    \centering
    \includegraphics[width=\linewidth]{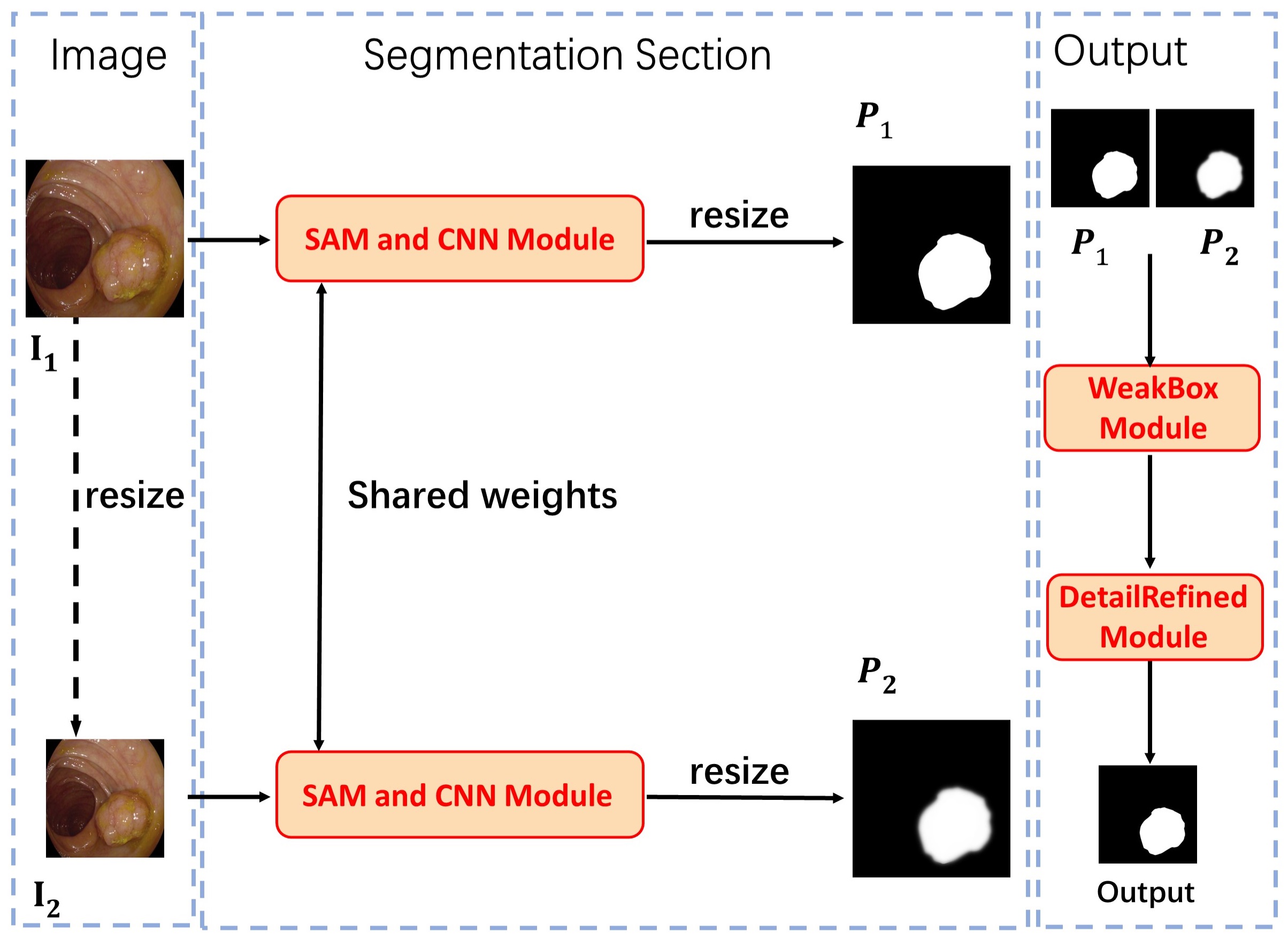}
    \caption{The pipeline for our proposed BiSeg-SAM. It includes: (1) an SAM and CNN Module that integrates local detail information into SAM; (2) an automatic box prompting mechanism via the WeakBox Module; and (3) the DetailRefine Module, aimed at learning clear edge information and richer image features.}
    \label{fig:network}
\end{figure}

The main contributions of this work can be summarized as follows:
\begin{itemize}
\item Introducing BiSeg-SAM, the weakly supervised method designed for SAM in medical binary classification segmentation, enabling automated prompting for the SAM model.

\item Enhancing bounding box generation and improving dataset generalization through MM2B transformation, which effectively addresses the transition between multiple and single foregrounds. This results in better segmentation performance for tasks involving multiple foreground objects or irregular and small foregrounds.

\item Integrating coarse learning and fine detail learning with a new training approach. This method initially captures main features and later focuses on boundaries and details, enhancing the model's ability to accurately segment complex and varied scenarios in binary segmentation tasks.
\end{itemize}

\section{Method}

Fig.~\ref{fig:network} 
depicts the entire pipeline of BiSeg-SAM, which consists of three modules. The input image is first resized and processed by the shared weights of the SAM and CNN Module to obtain segmentation results at different scales. These results are then fed into the WeakBox Module to generate coarse masks, which are further processed by the DetailRefine Module to produce the final output with enhanced boundary details. The technical details of these three modules are as follows:

\begin{figure}
    \centering
    \includegraphics[width=\linewidth]{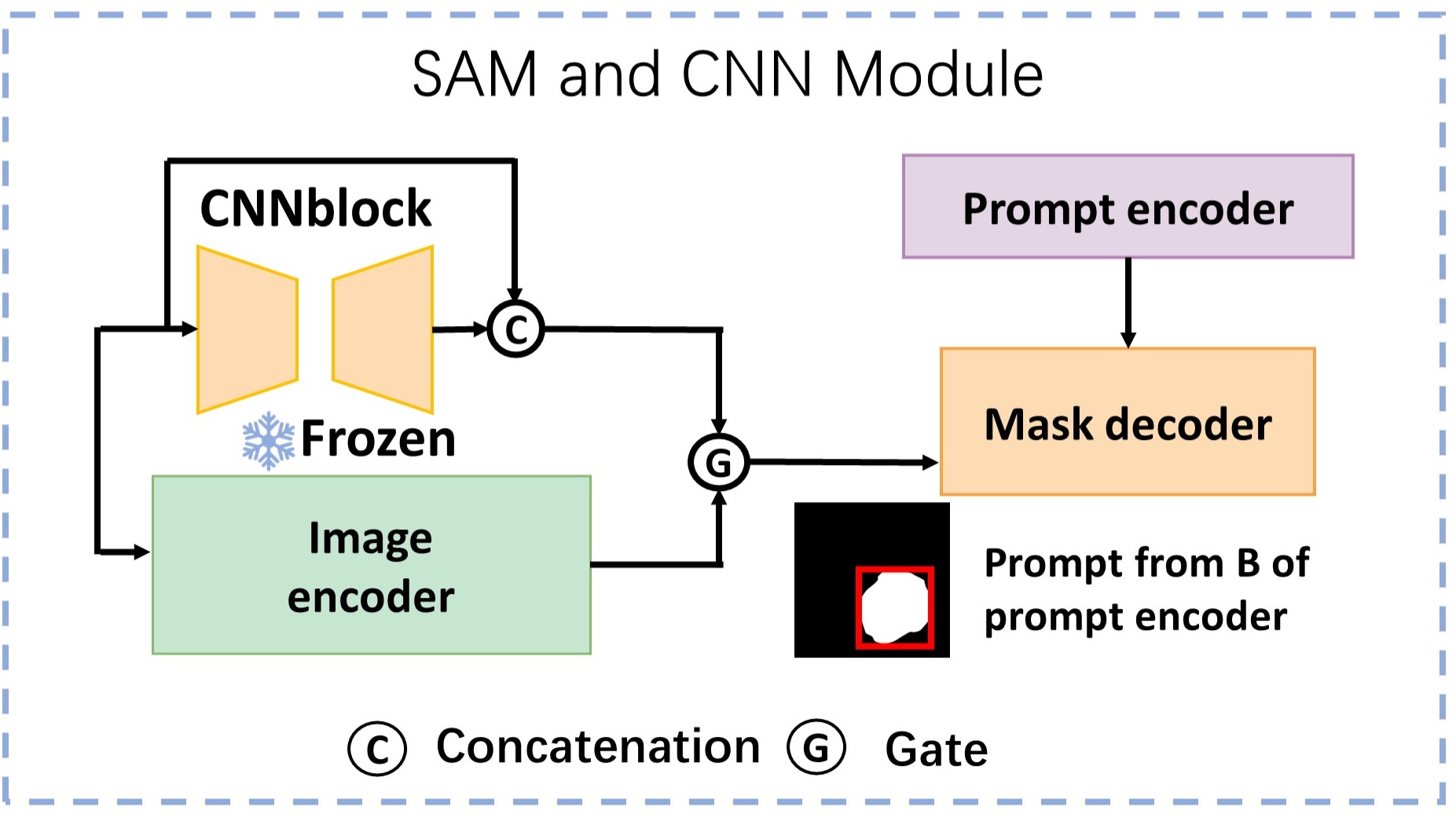}
    \caption{The architecture of the SAM and CNN Module. The image encoder from the SAM is frozen and integrated with a pre-trained CNN block. The output from the CNN block is concatenated with the frozen image encoder's output. The concatenated features are then passed through a gate before being fed into the mask decoder. Additionally, the bounding box generated during the MM2B block is used as a prompt for the B.}
    \label{fig:cnn}
\end{figure}

\subsection{Adaptively Global-Local Module}

This module integrates a CNN block structured on the principles of ResNet-18~\cite{b8}, designed to preprocess the input image and enhance localized features. To adapt the SAM model for medical image segmentation and address its limitations in extracting local features, we propose an improved CNN block. This module (see Fig.~\ref{fig:cnn}) not only extracts and concatenates local features from the images but also enhances the extraction and integration of these features by introducing a learnable weight coefficient $\alpha$~\cite{b11} to better balance the fusion of these two sets of features. Specifically, let the input image be $x \in \mathbb{R}^{H \times W \times C}$, where $H$ and $W$ represent the spatial dimensions, and $C$ denotes the number of channels. The input image is processed simultaneously by the CNN block and the SAM encoder. Feature concatenation and weight allocation are then utilized to obtain the feature map:

\begin{figure*}[t]
    \centering
    \includegraphics[width=0.75\linewidth]{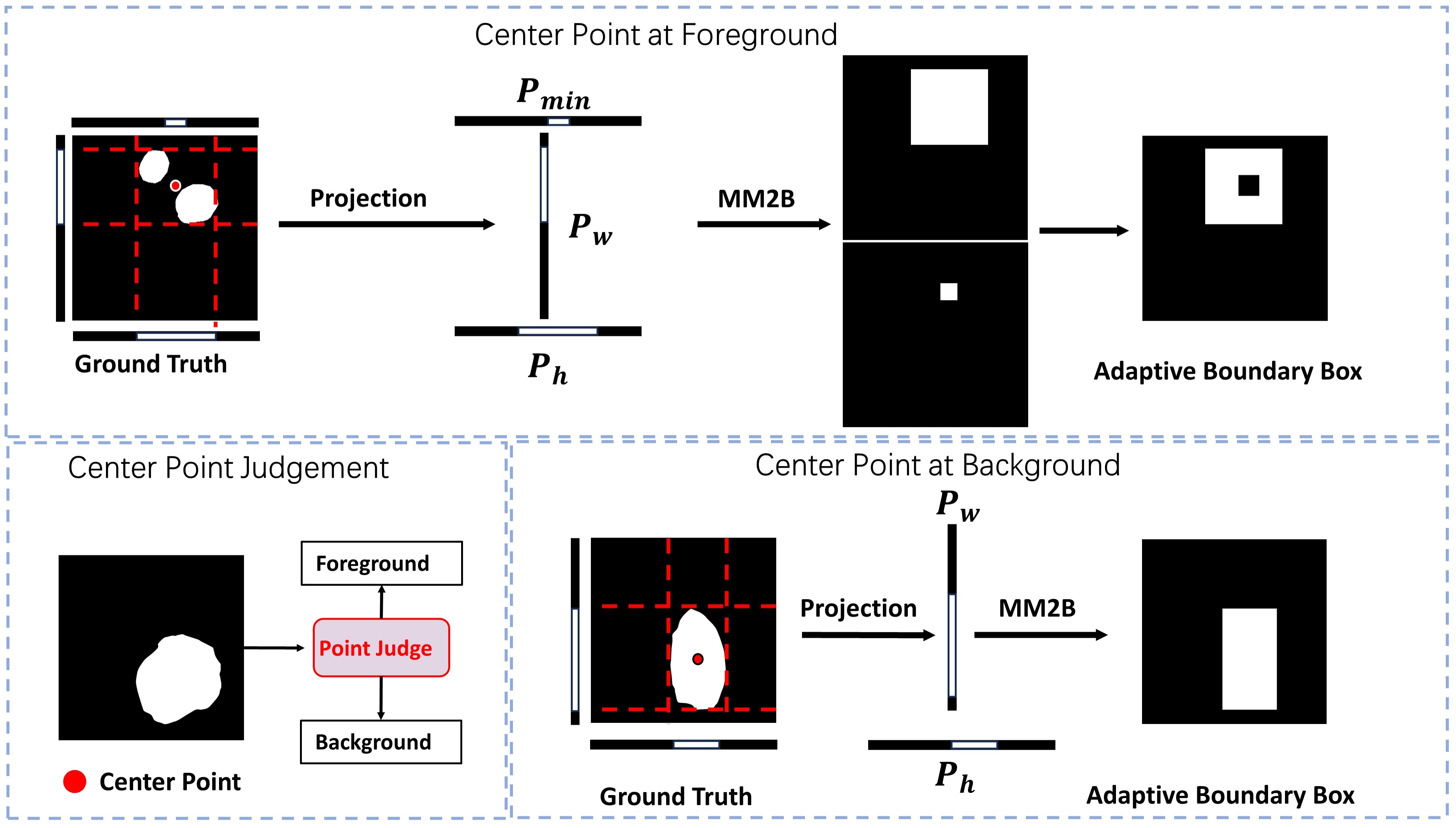}
    \caption{First, the center point is judged to determine whether it is in the foreground or background, clarifying the number of foreground objects. If it is a single foreground image, the bounding box is directly generated. If it is a multiple foreground image, the maximum and minimum values are used to generate a bounding box that encompasses all foregrounds.}
    \label{fig:centerpoint}
\end{figure*}

\begin{equation}
x = \alpha \cdot x_{\text{sam}} + (1 - \alpha) \cdot x_{\text{cnn}}.
\end{equation}
where $x_{\text{sam}}$ is the output of the image encoder of SAM, and $x_{\text{cnn}}$ is obtained by concatenating the processed features from the CNN block with the input image P1, P2.

\subsection{WeakBox Module}

This module, based on WeakPolyp~\cite{b9}, incorporates a Scale Consistency Loss and a Multi-choice Mask-to-Box (MM2B) Transformation. The MM2B method accurately captures the precise foreground area, thereby enhancing generalization.

\begin{figure}
    \centering
    \includegraphics[width=0.6\linewidth]{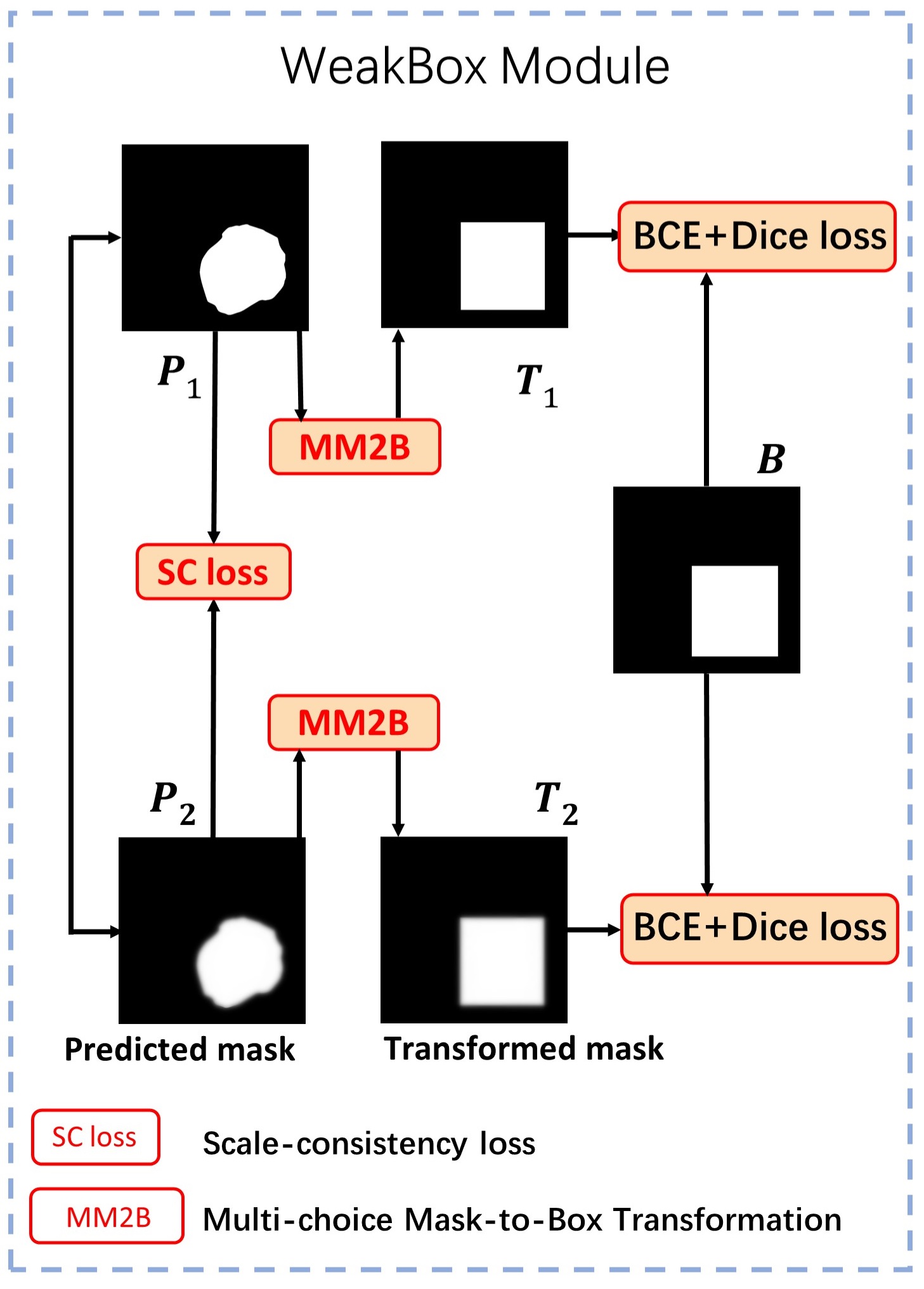}
    \caption{Technical details of the WeakBox Module. The multi-scale information $P_1$ and $P_2$ are partially optimized through SC loss. They are then transformed into bounding boxes via MM2B. The transformed bounding boxes $T_1$ and $T_2$ are compared with the ground truth bounding box $B$ using BCE+Dice loss. The final loss of this module is obtained by combining these two losses.}
    \label{fig:weak}
\end{figure}

\fakeparagraph{Multi-choice Mask-to-Box Transformation.}
In weakly supervised segmentation, bounding box annotations guide mask prediction, but precise edge definition is crucial for generalization. Bounding box bias may blur the target-background distinction. Adding a prompt decoder to SAM significantly boosts segmentation accuracy. Our approach (see Fig.~\ref{fig:weak}) enhances the weak box method with adaptive bounding boxes based on foreground center points, showing promising results.

\fakeparagraph{Projection Vector Generation and Target Localization.} 
Given the predicted mask \(P \in [0, 1]^{H \times W}\), projecting it into two vectors \(P_w\) and \(P_h\) can be represented as:

\begin{equation}
\begin{aligned}
P_w &= \max(P, \text{axis} = 0) \in [0, 1]^{1 \times W}, \\
P_h &= \max(P, \text{axis} = 1) \in [0, 1]^{H \times 1}.
\end{aligned}
\end{equation}

Here, $P_w$ and $P_h$ are obtained by applying max pooling to each column and row, representing projection vectors in the width and height directions.

\fakeparagraph{Bounding Box Generation Based on Center Point Status.}
Fig.~\ref{fig:centerpoint} and 
 Fig.~\ref{fig:weak} illustrate the process of generating bounding boxes based on the status of the center point within the mask $P \in [0, 1]^{H \times W}$. This novel approach aims to adaptively encapsulate the object of interest with minimal shape distortion, especially when dealing with multiple foreground objects simultaneously.

\fakeparagraph{Center Point at Foreground.}
When the center point is within the foreground (value $= 1$), the adaptive bounding box is determined through horizontal and vertical projections:

\begin{equation}
\begin{aligned}
    P'_w &= \text{repeat}(P_w, H, \text{axis} = 0) \in [0, 1]^{H \times W}, \\
    P'_h &= \text{repeat}(P_h, W, \text{axis} = 1) \in [0, 1]^{H \times W}, \\
    T_{\text{1}} &= \min(P'_w, P'_h) \in [0, 1]^{H \times W}.
\end{aligned}
\end{equation}

where the bounding box mask $T$ is constructed by back-projecting $P_w$ and $P_h$ into $P'_w$ and $P'_h$, respectively, and then taking the element-wise minimum. Simultaneously, we apply the $B$ as a prompt to the SAM model.

\fakeparagraph{Center Point at Background.}
In cases where the center point falls within the background (value $= 0$), our method generates two bounding boxes based on the longest and shortest diagonals to accommodate the object's extents accurately:

\emph{Maximum Boundary Box ($B_{\text{max}}$)}: 
\begin{equation}
\begin{aligned}
B_{\text{max}} &= \max(P'_w, P'_h) \in [0, 1]^{H \times W}.
\end{aligned}
\end{equation}

\emph{Minimum Boundary Box ($B_{\text{min}}$)}: 
To generate the minimum boundary box, we calculate the minimum distance between foreground pixels in the horizontal and vertical directions. Let $d_{h}$ and $d_{w}$ represent these minimum distances:

\begin{equation}
\begin{aligned}
d_{h} &= \min_{i, j} (|P(i, :) - P(j, :)|), \\
d_{w} &= \min_{i, j} (|P(:, i) - P(:, j)|).
\end{aligned}
\end{equation}

Then, the minimum boundary box $B_{\text{min}}$ can be defined as the bounding box that encapsulates these minimum distances:

\begin{equation}
\begin{aligned}
B_{\text{min}} &= \text{BoundingBox}(d_{h}, d_{w}) \in [0, 1]^{H \times W}.
\end{aligned}
\end{equation}

The final bounding box ($T_{\text{0}}$) is given by:

\begin{equation}
\begin{aligned}
T_{\text{0}} &= B_{\text{max}} - B_{\text{min}}.
\end{aligned}
\end{equation}

By using the maximum and minimum bounding boxes, our method generalizes effectively with coarse annotations. The final input to our model is the area defined by the maximum bounding box minus the minimum bounding box, ensuring all foreground objects are included. This approach not only ensures computational efficiency and simplicity but also enhances generalization by matching the sizes of the maximum and minimum bounding boxes, thereby integrating multiple foreground objects into a unified range. By focusing on the maximum and minimum bounding boxes, we can reduce complexity while maintaining accurate object encapsulation, which is particularly beneficial when dealing with multiple foreground objects simultaneously. This method reduces the risk of overfitting to specific object shapes and sizes in the training data, allowing the model to better generalize to unseen images with varying object configurations and complexities. Consequently, our approach improves the model's ability to handle weakly supervised data and enhances segmentation accuracy without requiring precise annotations.

\fakeparagraph{Foreground and Background Loss Components.}
For the foreground (value $= 1$) and background (value $= 0$) center points, we employ a combination of Binary Cross-Entropy (BCE) loss and Dice loss, specifically tailored to address the requirements of each condition:

\begin{equation}
\begin{aligned}
L_{\text{foreground}}(T, B) &= \frac{L_{\text{BCE}}(T1, B) + L_{\text{Dice}}(T1, B)}{2}, \\
L_{\text{background}}(T, B) &= \frac{L_{\text{BCE}}(T2, B) + L_{\text{Dice}}(T2, B)}{2}.
\end{aligned}
\end{equation}

where $T1$ and $T2$ denote the model's predictions for the foreground and background and $B$ represents the bounding box generated from coarse annotations of the dataset, respectively. The loss is chosen for their effectiveness in handling the binary nature of our task and their capacity to mitigate class imbalance issues.

\fakeparagraph{Adaptive Loss Function for Center Point Differentiation.}
Given the distinct characteristics of foreground and background center points in our model, we introduce an adaptive loss function that effectively incorporates this differentiation into the model training process. The total loss $L_{\text{Total}}$ is formulated as a weighted sum of the foreground and background losses, accommodating the nuances of each scenario:

\begin{equation}
L_{MM2B} = \beta L_{foreground}(T, B) + \gamma L_{background}(T, B). 
\end{equation}
where $T$ represents the predicted bounding box mask, while $B$ denotes the ground truth bounding box mask. The coefficients $\beta$ and $\gamma$ are empirically determined to balance the contributions of foreground and background losses, optimizing model performance by reflecting the relative importance of accurately classifying these areas.

\fakeparagraph{Automatic Prompt Integration for SAM.}
After determining the optimal bounding box, we automatically feed this bounding box as a prompt to the SAM model's prompt encoder. The integration leverages the bounding box generated by the WeakBox method to guide SAM in refining its segmentation predictions, ensuring that the segmented objects are accurately and efficiently encapsulated. The process enhances the model's performance by utilizing the bounding box as a precise input, thereby improving the overall segmentation accuracy and effectiveness in handling multiple foreground objects simultaneously.

\fakeparagraph{Scale Consistency (SC) Loss.}
To address prediction non-uniqueness in MM2B's sparse supervision, we apply the SC loss. This loss narrows the response gap between segmentation predictions obtained from different inputs of the same image, after resize and preprocessing. Here, $P_1$ corresponds to the segmentation prediction from input $I_1$, and $P_2$ corresponds to the segmentation prediction from input $I_2$. The SC loss concentrates on the segmented areas within bounding boxes to lessen prediction diversity:

\begin{equation}
L_{\text{SCloss}} = \frac{\sum_{(i,j) \in \text{box}} |P_{i,j}^1 - P_{i,j}^2|}{\sum_{(i,j) \in \text{box}} 1}.
\end{equation}
where $L_{\text{SCloss}}$ is the scale consistency loss, aimed at aligning the response values of $P_{i,j}^1$ and $P_{i,j}^2$ for each pixel $(i,j)$ in the bounding box to ensure consistent segmentation predictions.

\begin{figure}
    \centering
    \includegraphics[width=0.95\linewidth]{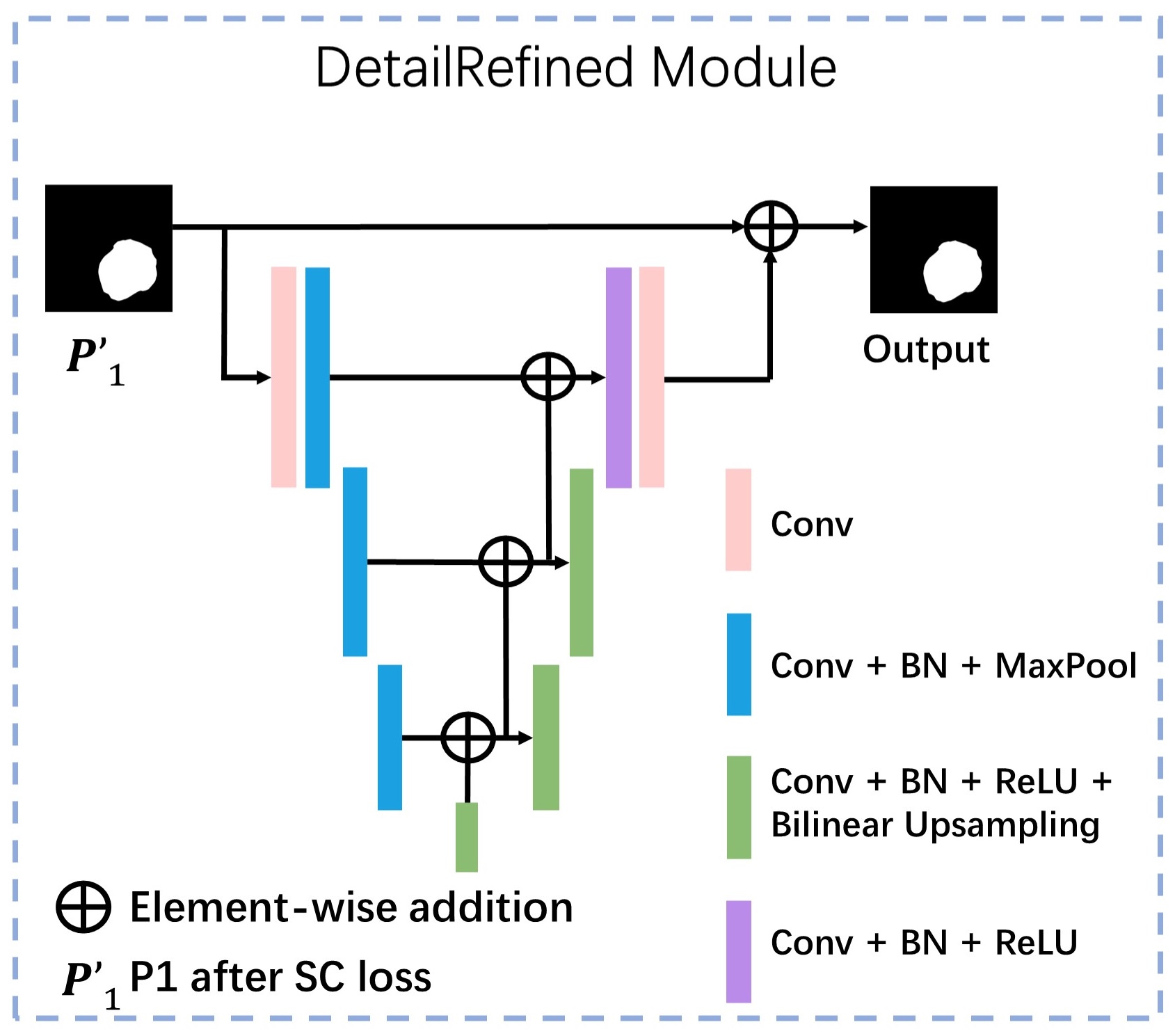}
    \caption{The architecture of the DetailRefine Module. It includes convolutional layers (Conv), batch normalization (BN), ReLU activation, max pooling (MaxPool), and bilinear upsampling layers. The module refines segmentation by combining coarse predictions with residual corrections through element-wise addition.}
    \label{fig:detail}
\end{figure}

\subsection{DetailRefine Module}

The DetailRefine Module (see Fig.~\ref{fig:detail}) is designed to refine segmentation tasks by enhancing the precision of object boundaries within medical images. This module improves segmentation accuracy by refining boundaries and structure recognition. One remarkable feature of this module is its capability to improve segmentation accuracy even in scenarios with data, thereby offering robustness and generalization to the model. Although it uses a small amount of GT masks to refine coarse segmentation results, it ensures the core framework of our method relies on weak supervision signals.

\fakeparagraph{Architecture.} This module consists of an encoder-decoder structure with skip connections, similar to the original U-Net. However, inspired by BasNet~\cite{b17}, it introduces residual blocks within both the encoding and decoding paths to capture more detailed features and facilitate the learning of fine-grained boundary information. The general formulation of the model's operation can be described by:

\begin{equation}
S_{\text{refined}} = S_{\text{coarse}} + S_{\text{residual}}.
\end{equation}

where $S_{\text{coarse}}$ represents the initial segmentation predictions, and $S_{\text{residual}}$ denotes the learned residual corrections. The final refined segmentation, $S_{\text{refined}}$, is obtained by adding the residual corrections to the initial predictions.

\fakeparagraph{Training Strategy.}
The DetailRefine Module is trained using a limited number of annotated samples from the same domain, specifically polyps and melanoma images. During the initial training phase, the model learns the coarse segmentation features. In the subsequent fine-tuning phase, the weights of the DetailRefine Module are frozen to prevent further updates. This strategy ensures that the model focuses on refining the segmentation boundaries without altering the learned features from the initial training phase.

\fakeparagraph{DetailRefine Loss.}
To train the DetailRefine Module, a composite loss function that combines the benefits of different loss metrics is utilized as follows:

\begin{equation}
L_{\text{DetailRefine}} = \lambda_1 L_{\text{Dice}}(S_{\text{refined}}, \text{GT}) + \lambda_2 L_{\text{CE}}(S_{\text{refined}}, \text{GT}).
\end{equation}

where $L_{\text{Dice}}$ and $L_{\text{CE}}$ represent the Dice and cross-entropy losses respectively, $S_{\text{refined}}$ is the refined segmentation output, and GT is the ground truth. $\lambda_1$ and $\lambda_2$ are two weighting coefficients used to adjust the contribution of $L_{\text{Dice}}$ and $L_{\text{CE}}$ within the DetailRefine loss function.

Even in scenarios where annotated data is limited, the DetailRefine Module effectively learns from the available annotated samples, leveraging the residual corrections to refine the segmentation predictions. This capability enhances the model's robustness and generalization, making it suitable for practical applications where fully annotated datasets may be scarce.

By freezing the weights during the fine-tuning phase, the module ensures that the initially learned features remain intact while focusing on improving the boundary details. This method allows for significant improvements in segmentation performance, particularly in terms of boundary precision, even with a limited number of training samples.

\subsection{Total Loss}
Utilizing the composite loss function mentioned above, BiSeg-SAM combines the strengths of the MM2B, SC, and DetailRefine losses:

\begin{equation}
L_{\text{total}} = L_{\text{MM2B}} + L_{\text{SCloss}} + L_{\text{DetailRefine}}.
\end{equation}

We only update the losses for the MM2B and SC modules and do not apply the DetailRefine loss during the main training process to ensure that our approach remains applicable to weak supervision. By using a limited number of ground truth masks to train a post-processing module suitable for binary classification tasks, we refine the segmentation details.

BiSeg-SAM maintains the original model structure, making it adaptable to various segmentation tasks. The loss components $L_{\text{MM2B}}$ and $L_{\text{SCloss}}$ are applied only during training, ensuring efficient inference without affecting speed. This design enhances segmentation accuracy, especially at object boundaries, and is robust across different imaging conditions.

\begin{table*}[t!]
\caption{Comparison of polyp segmentation methods on the multi-center datasets (Kvasir, CVC-ClinicDB, CVC-ColonDB, Endoscene, and ETIS). Bold values indicate the highest scores in each column. It can be seen that, compared to other fine-tuned SAM models, our network performs better in tasks involving multiple foreground objects.}
\begin{center}
\begin{tabular}{|c|c|cc|cc|cc|cc|cc|}
\hline
\multirow{2}{*}{\centering\textbf{Models}} & \multirow{2}{*}{\centering\textbf{Year}} & \multicolumn{2}{|c|}{\textbf{Kvasir}} & \multicolumn{2}{|c|}{\textbf{CVC-ClinicDB}} & \multicolumn{2}{|c|}{\textbf{CVC-ColonDB}} & \multicolumn{2}{|c|}{\textbf{Endoscene}} & \multicolumn{2}{|c|}{\textbf{ETIS}} \\
\cline{3-12} 
 &  & \textbf{\textit{DSC}} & \textbf{\textit{mIoU}} & \textbf{\textit{DSC}} & \textbf{\textit{mIoU}} & \textbf{\textit{DSC}} & \textbf{\textit{mIoU}} & \textbf{\textit{DSC}} & \textbf{\textit{mIoU}} & \textbf{\textit{DSC}} & \textbf{\textit{mIoU}} \\
\hline

U-Net~\cite{b18} & 2015 & 0.818 & 0.746 & 0.823 & 0.755 & 0.512 & 0.444 & 0.710 & 0.627 & 0.398 & 0.335 \\
UNet++~\cite{b4} & 2018 & 0.821 & 0.743 & 0.794 & 0.729 & 0.483 & 0.410 & 0.707 & 0.624 & 0.401 & 0.344 \\
PraNet~\cite{b6} & 2020 & 0.898 & 0.840 & 0.899 & 0.849 & 0.712 & 0.640 & 0.871 & 0.797 & 0.628 & 0.567 \\
Polyp-PVT~\cite{b5} & 2021 & 0.917 & 0.864 & \textbf{0.937} & \textbf{0.889} & 0.808 & 0.727 & 0.900 & 0.833 & 0.787 & 0.706 \\
MSNet~\cite{b14} & 2021 & 0.905 & 0.849 & 0.918 & 0.869 & 0.751 & 0.671 & 0.865 & 0.799 & 0.723 & 0.652 \\
Swin-Unet~\cite{b20} & 2022 & 0.867 & 0.851 & 0.910 & 0.883 & 0.868 & 0.815 & 0.811 & 0.807 & 0.741 & 0.660 \\
polyp-SAM~\cite{b12} & 2023 & 0.902 & 0.863 & 0.921 & 0.877 & \textbf{0.894} & 0.843 & \textbf{0.924} & \textbf{0.882} & 0.903 & \textbf{0.852} \\
WeakPolyp~\cite{b9} & 2023 & 0.878 & 0.815 & 0.863 & 0.794 & 0.766 & 0.676 & 0.866 & 0.790 & 0.678 & 0.604 \\
SAMed~\cite{b29} & 2023 & 0.842 & 0.836 & 0.887 & 0.851 & 0.792 & 0.774 & 0.864 & 0.787 & 0.764 & 0.683 \\
\hline
BiSeg-SAM & 2024 & \textbf{0.919} & \textbf{0.882} & 0.923 & 0.874 & 0.887 & \textbf{0.862} & 0.907 & 0.865 & \textbf{0.904} & 0.844 \\
\hline
\end{tabular}
\label{tab:polyp_metrics}
\end{center}
\end{table*}

BiSeg-SAM maintains the original model structure, making it adaptable to various segmentation tasks. The loss components $L_{\text{MM2B}}$ and $L_{\text{SCloss}}$ are applied only during training, ensuring efficient inference without affecting speed. The design enhances segmentation accuracy, especially at object boundaries, and is robust across different imaging conditions.

\begin{figure*}[t!]
    \centering
    \includegraphics[width=0.79\linewidth]{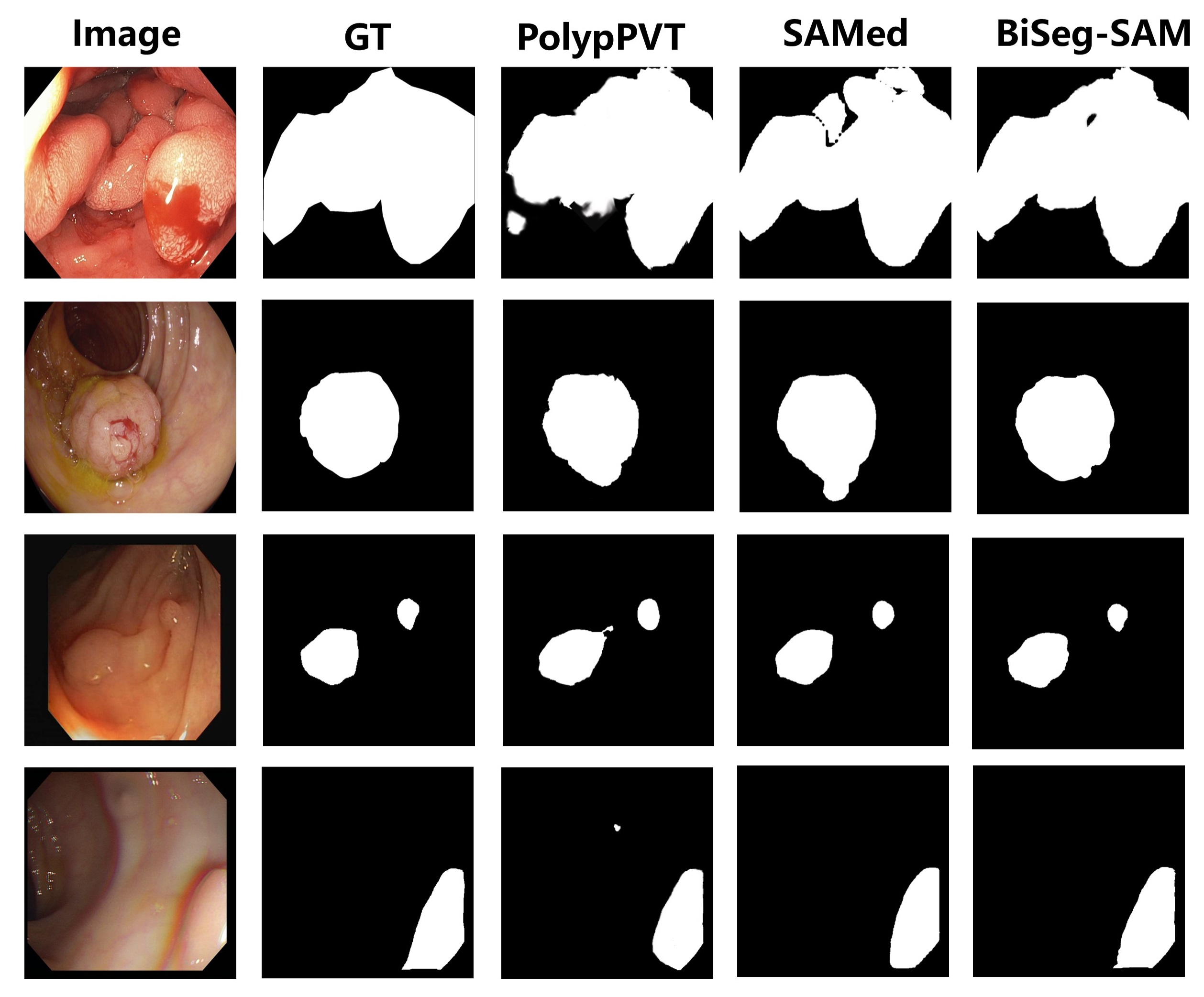}
    \caption{The qualitative comparisons between BiSeg-SAM and several other approaches. Our network demonstrates clearer edge information for single foreground polyps and better differentiation in multi-foreground scenarios. It also effectively eliminates many noise artifacts, resulting in more accurate contour delineation compared to using SAM alone on medical images.}
    \label{fig:visual}
\end{figure*}

\section{Experiments}
\subsection{Dataset}
To assess the performance of our model in binary segmentation tasks, we utilized two primary datasets:
1. Polyp datasets: This evaluation encompasses five widely recognized polyp segmentation datasets, namely Kvasir~\cite{b24}, CVC-ClinicDB~\cite{b25}, CVC-ColonDB~\cite{b26}, EndoScene~\cite{b28}, and ETIS~\cite{b27}. Consistency with previous benchmarks is maintained by adhering to the dataset division protocol established in Polyp-PVT.
2. ISIC17 datasets: The ISIC dataset comprises dermatological lesion images in high resolution, accompanied by detailed annotations for both segmentation and classification tasks. Specifically, the ISIC2017 challenge~\cite{isic} provided 2000 lesion images along with their precise ground truth delineations.

Additionally, to validate the effectiveness of the CNN block and DetailRefine module in our model, we employed an additional multi-class dataset, Synapse\footnote{\href{https://www.synapse.org/!Synapse:syn3193805/wiki/217789}{https://www.synapse.org/!Synapse:syn3193805/wiki/217789}}, for multi-organ CT segmentation. This dataset comprises 30 cases of abdominal CT scans. Following the split protocol used in SwinUnet~\cite{b20}, using an 18:12 training-testing split ratio, we allocated 18 cases for the training set and the remaining 12 cases for the testing set. We evaluated the model's performance on eight abdominal organs—namely, the aorta, gallbladder, spleen, left kidney, right kidney, liver, pancreas, and stomach—using the 95\% Hausdorff Distance (HD95) and Dice Score (DSC) metrics. Detailed preprocessing steps, such as intensity normalization and data augmentation including random rotations and flips, were applied to ensure robust training.

\subsection{Implementation Details}

For the implementation of BiSeg-SAM, we provide additional details to ensure reproducibility and clarity. The model architecture, implemented using the "vit\_b" variant of the SAM architecture, is composed of three main modules: the Adaptively Global-Local Module, the WeakBox Module, and the DetailRefine Module. Each module is equipped with specific configurations and parameters to optimize performance. We utilize a uniform input size of 512x512 pixels for all images, and a standard data augmentation pipeline is applied, including random flipping, random rotation, and multi-scale training, aligning with strategies employed in SAMed~\cite{b29}.

During training, we configure the loss function weights to assign 0.2 weight to the cross-entropy loss and 0.8 weight to the Dice loss, based on empirical observations of their relative importance. The model is trained end-to-end using the AdamW optimizer, with an initial learning rate set to 1e-4 and a batch size of 16. Training is conducted over 200 epochs to ensure comprehensive learning and parameter optimization. All experiments are performed using 2 NVIDIA RTX A5500 GPUs for efficient computation.

\begin{table}[t!]
\caption{Comparisons of different methods on the ISIC17 dataset. Bold values indicate the highest scores in each column. It can be seen that our network performs exceptionally well in segmentation when the foreground objects occupy a significant portion of the image.}
\begin{center}
\resizebox{\linewidth}{!}{ 
\begin{tabular}{|c|c|c|c|c|c|}
\hline
\textbf{Model} & \textbf{\textit{mIoU}} & \textbf{\textit{DSC}} & \textbf{\textit{Acc}} & \textbf{\textit{Sen}} & \textbf{\textit{Spe}}\\
\hline
UNet~\cite{b18} & 0.7698 & 0.8699 & 0.9565 & 0.8682 & 0.9743\\
UNetV2~\cite{b30} & 0.7735 & 0.8723 & 0.9584 & 0.8485 & 0.9805\\
Swin-Unet~\cite{b20} & 0.8089 & 0.8199 & 0.9476 & 0.8806 & 0.9605\\
TransFuse~\cite{b21} & 0.7921 & \textbf{0.8840} & 0.9617 & 0.8714 & 0.9768\\
MALUNet~\cite{b22} & 0.7878 & 0.8813 & 0.9618 & 0.8478 & \textbf{0.9847}\\
SAMed~\cite{b29} & 0.8087 & 0.8739 & 0.9589 & 0.8890 & 0.9762\\
\hline
BiSeg-SAM & \textbf{0.8159} & 0.8832 & \textbf{0.9674} & \textbf{0.8894} & 0.9792 \\
\hline

\end{tabular}
}
\label{table:models}
\end{center}
\end{table}

\begin{table}[t!]
\caption{Multi-center datasets average results of ablation study on applying multi-scale prediction.}
\begin{center}
\resizebox{\linewidth}{!}{
\begin{tabular}{|c|c|c|c|c|c|}
\hline
\textbf{SAM} & \textbf{CNN} & \textbf{WeakBox} & \textbf{DetailRefine} & \textbf{\textit{Avg. DSC }} & \textbf{\textit{Avg. mIoU }} \\
\hline
\checkmark & & & & 0.759 & 0.724 \\
\checkmark & \checkmark & & & 0.776 & 0.755 \\
\checkmark & & \checkmark & & 0.802 & 0.791 \\
\checkmark & & & \checkmark & 0.817 & 0.787 \\
\checkmark & \checkmark & & \checkmark & 0.824 & 0.801 \\
\checkmark & \checkmark & \checkmark & & 0.839 & 0.822 \\
\checkmark & \checkmark & \checkmark & \checkmark & \textbf{0.846} & \textbf{0.827} \\
\hline
\end{tabular}
}
\label{tab:polyp_ablation}
\end{center}
\end{table}

\begin{table}[t!]
\caption{Ablation study results on Synapse multi-organ CT segmentation dataset. The metrics include Dice Score (DSC) and 95\% Hausdorff Distance (HD95) for each organ.}
\begin{center}
\resizebox{\linewidth}{!}{
\begin{tabular}{|c|c|c|c|c|}
\hline
\textbf{SAM} & \textbf{CNN} & \textbf{DetailRefine} & \textbf{\textit{Avg. DSC }} & \textbf{\textit{Avg. HD95 (mm)}} \\
\hline
\checkmark & & & 0.7963 & 22.5 \\
\checkmark & \checkmark & & 0.8002 & 22.1 \\
\checkmark & & \checkmark & 0.8232 & 20.6 \\
\checkmark & \checkmark & \checkmark & \textbf{0.8267} & \textbf{20.2} \\
\hline

\end{tabular}
}
\label{tab:synapse_ablation}
\end{center}
\end{table}

\subsection{Comparison with SOTAs}

In our comparative analysis against state-of-the-art (SOTA) methods, BiSeg-SAM demonstrates superior performance across multiple multicenter polyp datasets (Kvasir, CVC-ClinicDB, CVC-ColonDB, Endoscene, ETIS) (see Table~\ref{tab:polyp_metrics}). Specifically, on the Kvasir dataset, BiSeg-SAM achieves the highest Dice Similarity Coefficient (DSC) and mean Intersection over Union (mIoU) scores of \textbf{0.919} and \textbf{0.882}, respectively, surpassing all compared models. Additionally, on other datasets like CVC-ColonDB and ETIS, BiSeg-SAM also shows competitive performance, particularly excelling in tasks that involve multiple foreground objects.

For the ISIC17 skin lesion segmentation dataset, BiSeg-SAM achieves the highest mIoU (\textbf{0.8159}) and accuracy (\textbf{0.9674}) scores, and also records the best sensitivity (\textbf{0.8894}) among all compared methods (see Table~\ref{table:models}). These results indicate that BiSeg-SAM is particularly effective in segmentation tasks where the foreground objects occupy a significant portion of the image, accurately delineating the target regions while maintaining high generalization performance.

\subsection{Ablation Study}

The ablation study provides insights into the individual contributions of each component in BiSeg-SAM.

\begin{itemize}
    \item \textbf{Adaptively Global-Local Module} and \textbf{WeakBox Module} both contribute significantly to the improvement in segmentation performance. The ablation study on multicenter datasets (see Table~\ref{tab:polyp_ablation}) shows that the baseline SAM module achieves an average DSC of \textbf{0.759}. Introducing the WeakBox Module leads to a substantial increase in DSC to \textbf{0.802}, demonstrating the effectiveness of guided learning with weak annotations.

    \item The \textbf{DetailRefine Module} plays a crucial role in refining boundary features. Adding this module improves the DSC to \textbf{0.817}, and when combined with the CNN block, the DSC further increases to \textbf{0.824}. This highlights the importance of boundary refinement in enhancing segmentation accuracy.

    \item The complete model, incorporating all modules, achieves the best results across all evaluation metrics, with an average DSC of \textbf{0.846} and an average mIoU of \textbf{0.827}. This demonstrates the synergistic effect of the components in improving both segmentation accuracy and robustness.
\end{itemize}

The ablation study results on the Synapse multi-organ CT segmentation dataset (see Table~\ref{tab:synapse_ablation}) also demonstrate the contributions of the CNN block and DetailRefine Module. Specifically, incorporating the DetailRefine Module leads to an increase in the average Dice score from \textbf{0.7963} to \textbf{0.8232}, and a reduction in the 95\% Hausdorff Distance (HD95) from \textbf{22.5 mm} to \textbf{20.6 mm}, indicating the effectiveness of boundary feature refinement in enhancing segmentation performance. The final combination of all modules achieves an average DSC of \textbf{0.8267} and HD95 of \textbf{20.2 mm}, underscoring the crucial role of local feature extraction and boundary refinement in improving segmentation accuracy and robustness.

These results clearly illustrate that each component of the BiSeg-SAM model makes significant contributions to enhancing segmentation performance. Notably, the refinement of detailed features under weak supervision allows the model to achieve high accuracy and robustness across diverse datasets and imaging conditions.

\section{Conclusion}

In this study, we introduced BiSeg-SAM, a novel weakly supervised post-processing framework to enhance binary segmentation in SAM for medical image analysis. BiSeg-SAM integrates adaptive bounding box generation, weak supervision, and detailed boundary refinement techniques to address the limitations of existing models in handling complex medical images with limited annotations. Experimental results show that BiSeg-SAM achieves superior performance compared to SOTA methods across multiple datasets, including polyp and skin cancer segmentation tasks.

The comprehensive approach of BiSeg-SAM not only reduces the dependency on extensive pixel-level annotations but also ensures high segmentation accuracy, making it a practical solution for medical image analysis. Future work will extend this framework to other medical imaging modalities, such as MRI and ultrasound, to assess its versatility and robustness. Additionally, we plan to explore the integration of BiSeg-SAM with other advanced imaging techniques, like multi-modal fusion and generative models, as well as machine learning models, to further improve its applicability and efficiency.

\bibliographystyle{splncs04}
\bibliography{egbib}

\end{document}